\documentclass[letterpaper, 10 pt, journal, twoside]{IEEEtran}

\usepackage{hyperref, url, amsmath, amssymb, booktabs, graphicx, bm, color, algorithm, tabularx, enumitem, multirow, threeparttable, flushend, cite}
\usepackage[noend]{algpseudocode}
\usepackage{amsfonts}
\usepackage{booktabs} 
\usepackage{multirow} 
\usepackage{makecell} 
\usepackage{float}
\usepackage{algpseudocode}
\newcommand{\vg}[1]{\bm{#1}}
\renewcommand{\v}[1]{\mathbf{#1}}
\hypersetup{
    colorlinks=true,
    linkcolor=blue,
    citecolor=blue,
    filecolor=magenta,      
    urlcolor=cyan,
    pdftitle={Overleaf Example},
    pdfpagemode=FullScreen,
    }
\graphicspath{{images/}}
\usepackage{soul}
\usepackage{cuted}
\usepackage{capt-of}

\usepackage{microtype}

\usepackage[textsize=tiny]{todonotes}
\usepackage{ifthen}
\newboolean{comments}
\setboolean{comments}{true} 

\ifthenelse{\boolean{comments}}{
  \setlength{\marginparwidth}{1.5cm}
  \newcommand{\zkn}[1]{\todo[color=orange!20, size=\tiny]{Zsolt: #1}}
  \newcommand{\zk}[1]{\textcolor{blue}{Zsolt: #1}}
}{
  \renewcommand{\zk}[1]{}
  \renewcommand{\zkn}[1]{}
}

\begin{document}
\bstctlcite{IEEEexample:BSTcontrol} 

\title{\LARGE Opt2VLA: 
Force-Aware Vision-Language-Action for Contact-Rich Humanoid Whole-Body Manipulation
}

\author{
Fukang Liu\textsuperscript{*},
Yipu Chen\textsuperscript{*},
Jaehwi Jang,
Danfei Xu,
Zsolt Kira, and
Ye Zhao%

\thanks{\textsuperscript{*} Equal contribution. Authors are with the Institute for Robotics and Intelligent Machines, Georgia Institute of Technology, Atlanta, GA, USA.}
}

\maketitle
\thispagestyle{empty}
\pagestyle{empty}

\begin{strip}
\vspace{-1.5cm}
    \centering
      \includegraphics[width=1\linewidth]{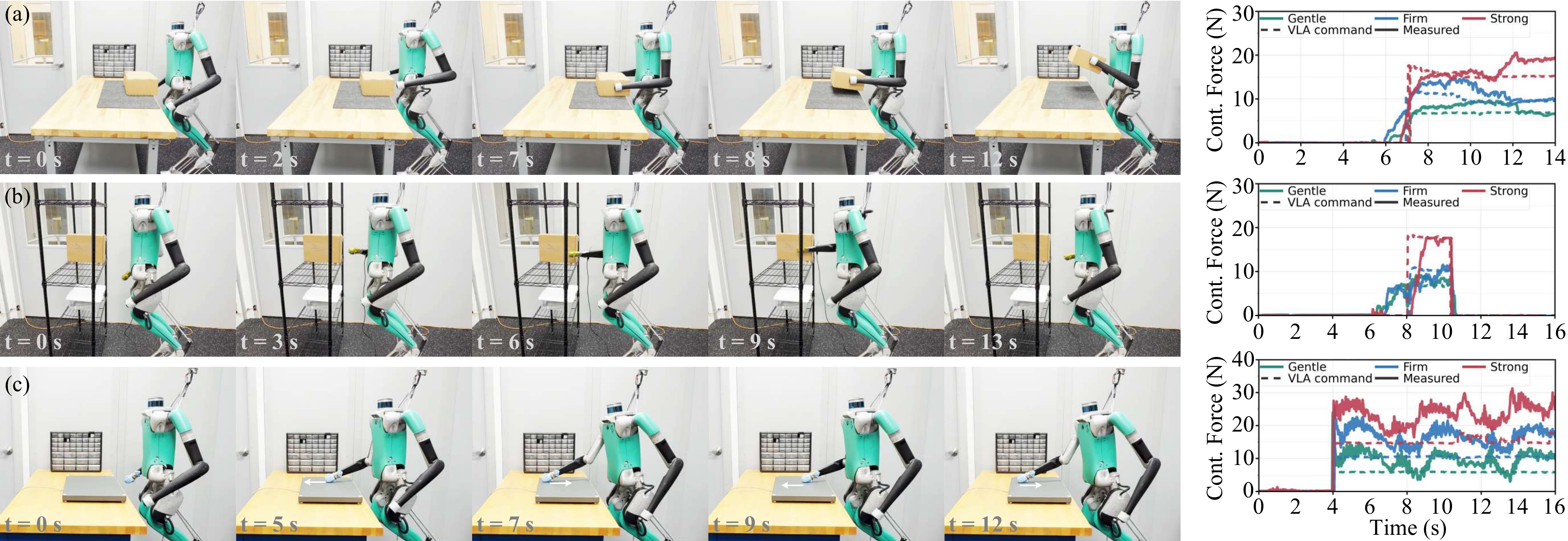}
      \captionof{figure}{
      \textbf{Opt2VLA enables autonomous, contact-rich whole-body humanoid manipulation with language-specified force levels.}
       (a) box pickup, (b) shelf-box pushing against a designated shelf wall, and (c) surface wiping. Snapshots show the task progression. Plots compare VLA-issued force commands (dashed) and measured contact forces (solid) for selected hardware trials under \textit{gentle}, \textit{firm}, and \textit{strong} force prompts. Opt2VLA jointly predicts motion and contact-force commands, enabling language-conditioned force modulation during autonomous whole-body manipulation.
       }
      \label{fig:hardware_overview}
\end{strip}

\begin{abstract}
Humanoid robots are expected to perform diverse human-level tasks in daily environments, many of which require precise regulation of interaction forces. While recent vision-language-action (VLA) models have shown promise for semantic planning and visuomotor control, existing humanoid systems primarily represent actions through geometric motion goals and rely on whole-body controllers focused on motion tracking, with limited explicit reasoning or control of interaction forces.
This limitation is particularly relevant in contact-rich tasks, where geometrically similar motions may require different force regimes depending on the task context and where visual observations may become unreliable after contact.
In this work, we present Opt2VLA, a force-aware VLA framework that introduces explicit force commands at the VLA-to-control interface for humanoid whole-body manipulation. A single multi-task VLA policy
jointly predicts both geometric motion goals and continuous contact-force references, which are tracked by task-specific reinforcement learning (RL)-based whole-body controllers. To provide scalable and physically grounded supervision, we generate dynamically feasible and contact-consistent training data via whole-body trajectory optimization (TO) with explicit force references. 
We evaluate Opt2VLA on three contact-rich humanoid tasks and show that explicit force conditioning enables more accurate and consistent force regulation than motion-only control, while physically grounded torque supervision from TO further improves force tracking accuracy and stability. Closed-loop evaluations further demonstrate language-conditioned force modulation with Opt2VLA in simulation and on humanoid hardware.
\url{https://opt2vla.github.io}

\end{abstract}

\section{Introduction}
Humanoid robots are designed to operate in human-centered environments, where many everyday tasks require sustained physical interaction with the world~\cite{barreiros2025learning, gu2025humanoidsurvey}. Beyond geometric accuracy, such interactions demand precise regulation of contact forces to ensure safety, stability, and task effectiveness, for example when pushing objects into constrained spaces, maintaining contact while sliding along surfaces, or interacting under partial visual occlusion. As humanoid systems move beyond scripted motions toward autonomous operation, the ability to reason about and control interaction forces becomes increasingly central to robust whole-body behavior.

Recent progress in whole-body control (WBC) has enabled robust humanoid behaviors across a wide range of locomotion and manipulation tasks, demonstrating impressive performance in the real world~\cite{gu2025humanoidsurvey, khazoom2024tailoring, humanplus}. These advances provide a strong foundation for humanoid low-level execution.
In parallel, vision-language-action (VLA) models have shown promise in enabling robots to interpret natural language instructions and plan long-horizon behaviors from visual observations~\cite{kim2024openvla, zitkovich2023rt, gr00tn1_2025, intelligence2025pi}. Recent humanoid systems have begun integrating these capabilities with whole-body control, showing progress toward instruction-driven whole-body behaviors~\cite{gr00tn1_2025,ding2025humanoid,yuan2025being,jiang2025wholebodyvla,luo2026sonic}.
However, these approaches primarily use motion-based action representations, such as end-effector poses or joint-position targets, with limited emphasis on explicitly specifying desired interaction forces at the VLA-to-control interface.

Despite the growing success of humanoid VLA models, this motion-based abstraction can be insufficient for contact-rich tasks.
Geometric motion goals alone may not fully specify the intended interaction, since similar motions may require different levels of contact forces depending on the task context, object properties, or safety constraints. Recent contact-aware and tactile-based VLA methods have begun incorporating force or tactile feedback into manipulation or loco-manipulation~\cite{yu2025forcevla, zhang2025ta, zhang2026fwbc}. However, these approaches generally use contact signals as observations or feedback for action prediction, rather than explicitly specifying desired interaction force as a high-level reference jointly planned with motion.
Incorporating force into a humanoid autonomy pipeline, however, presents additional challenges. At the high-level, force references should remain compact enough for VLA prediction while providing sufficient information for robust whole-body execution under uncertain contact conditions.
Learning to predict such references also requires structured supervision that aligns motion goals and force references with visual observations and language instructions.
Obtaining this supervision is particularly challenging for humanoids, for which collecting contact-rich demonstrations through teleoperation requires substantial effort~\cite{yuan2025being,gr00tn1_2025}, and providing diverse and consistently labeled force references adds further requirements to the data collection process.

These observations motivate an explicit motion-force interface in humanoid VLA systems. We separate task-level specification of the of the desired interaction force from its joint-level realization by whole-body control. Although force can also be modulated indirectly through motion commands~\cite{rouxel2024multi}, we expose the desired contact force explicitly alongside geometric motion goals, so that the high-level policy can specify both where to move and how strongly to interact.

\begin{figure*}[t!]
\centering
\includegraphics[width=1\linewidth]{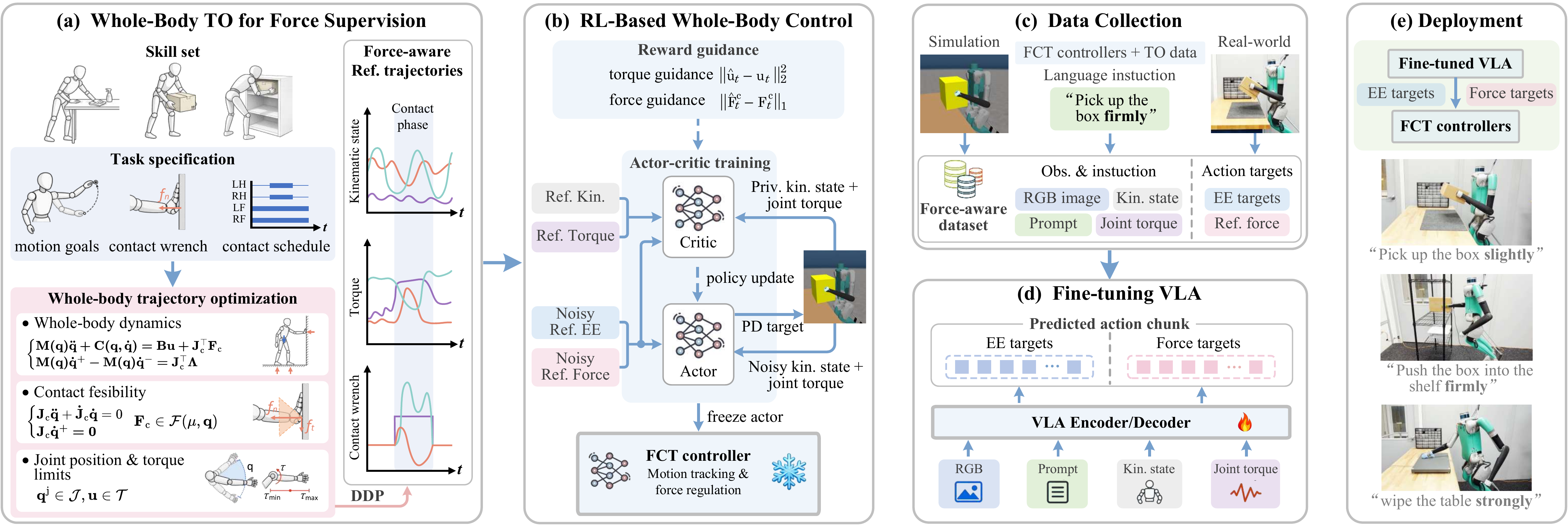}
\vspace{-0.2cm}
\caption{\textbf{Overview of Opt2VLA}. 
(a) Whole-body TO generates physically consistent motion, joint torque, and contact force trajectories under full-body dynamics and contact constraints. 
(b) Force-conditioned controller with torque supervision (FCT) is trained to track geometric motion goals (i.e., end-effector $3$D target positions) while regulating interaction force, using privileged torque supervision only during training.
(c) Large-scale simulation rollouts, together with a smaller set of real-world executions, are used to collect multimodal data including vision, language prompts, robot kinematic state, measured joint torques, end effector targets, and target-force labels.
(d) A pretrained vision–language–action model is then fine-tuned to predict motion goals and contact-force references from language, vision, kinematic state, and joint torque, and (e) deployed to execute force-aware behaviors through the FCT controllers. 
}
\label{fig:method}
\vspace{-0.2cm}
\end{figure*}

In this work, we present \textbf{Opt2VLA}, a force-aware VLA framework for humanoid whole-body control, as shown in Fig.~\ref{fig:method}. Opt2VLA is built around two complementary ideas. First, a single multi-task VLA policy, instantiated from a pretrained VLA model~\cite{gr00tn1_2025},  jointly predicts geometric motion goals and continuous contact-force references from task instructions, visual observations, robot states, and measured joint-torque feedback.  
Second, we use task-specific force-conditioned control with torque supervision (FCT), where TO-derived joint-torque references provide training-time guidance to improve force realization. Rollouts of these controllers further provide multimodal data paired with motion, force, and language targets for VLA fine-tuning, without requiring large-scale teleoperated data collection. This hierarchy decouples task-level motion-force prediction from low-level execution while allowing the predicted force commands to directly condition whole-body control. 

We evaluate the proposed framework on a set of contact-rich humanoid tasks that require precise force regulation. The results show improved force regulation over the evaluated motion-tracking baseline, while physically grounded torque supervision from TO further enhances force tracking accuracy and stability. Our contributions are:
\begin{itemize}
    \item We formulate force-aware VLA control for humanoid whole-body systems through an explicit interface of motion and contact-force commands.
    \item We present a hierarchical framework that integrates vision-language planning with force-aware RL-based whole-body control, using TO-generated references for controller training and controller rollouts for VLA fine-tuning.
    \item We demonstrate improved force regulation through explicit force conditioning and TO-derived torque supervision, and validate the framework through sim-to-real transfer on contact-rich humanoid tasks. We will publicly release the force-aware humanoid dataset collected through rollouts of controllers trained using TO-generated references.
\end{itemize}

\section{Related Work}
\subsection{Humanoid Whole-Body Control}
Model-based control and optimization methods that rely on accurate dynamic models have long served as a foundation for humanoid whole-body behaviors~\cite{kajita2003biped, wensing2023optimization, koenemann2015whole, khazoom2024tailoring, adu2023exploring}. However, their dependence on precise model fidelity and contact assumptions makes real-time deployment for complex humanoid interactions challenging under uncertainty in robot dynamics and environment contacts. 
Whole-body force regulation has also been demonstrated on position-controlled humanoids~\cite{rouxel2024multi}.

RL-based approaches, including motion imitation, have emerged as prominent methods for humanoid whole-body control, leveraging large-scale simulation and task-specific objectives to acquire diverse behaviors~\cite{peng2018deepmimic, radosavovic2024real, dao2024sim}. A common paradigm guides humanoid motion using expert trajectories retargeted from human demonstrations~\cite{cheng2024expressive, dugar2025learning, humanplus, he2024learning}. While such pipelines have enabled versatile motions on humanoid hardware, the embodiment gap between humans and humanoids makes the retargeting challenging, and the resulting trajectories are not always kinematically or dynamically feasible, which can degrade data quality. Moreover, the absence of explicit force and torque supervision in these motion priors limits their ability to provide physically grounded supervision for contact-rich interactions requiring precise force regulation. Relatedly, SoftMimic~\cite{margolis2025softmimic} augments motion imitation with compliant reference trajectories to enable stiffness-modulated whole-body behaviors under external disturbances, but does not explicitly represent or command task-level interaction forces.
To address these challenges, recent work has explored combining model-based optimization with RL for legged-robot control~\cite{jenelten2024dtc, fuchioka2022opt, dai2026walk, li2026accelerating}. TO provides dynamically feasible references by incorporating robot dynamics and contact constraints, and has been increasingly leveraged to guide learning-based locomotion and whole-body control~\cite{batke2022optimizing, marew2024biomechanics, chaikovskaya2023benchmarking, krishna2022linear,olkin2025chasing}.
More recently, Opt2Skill~\cite{liu2025opt2skill} leverages full-order dynamic TO for humanoid loco-manipulation, showing that dynamically feasible motion and torque references improve motion tracking and contact-rich execution. However, these methods are primarily conditioned on geometric motion references and do not expose desired contact force as an independent task-level command. As a result, different interaction-force requirements cannot be specified directly from high-level task instructions without modifying the motion reference or controller.

\subsection{Vision-Language-Action for Humanoids}
Recent advances in VLA models demonstrate that large-scale multimodal pretraining can endow robots with strong generalization and language grounding~\cite{kim2024openvla, zitkovich2023rt}, motivating their extension to humanoid autonomy~\cite{xu2024humanvla}. Recent humanoid-focused VLA systems ground language and perception into whole-body actions, typically represented as kinematic targets such as end-effector poses or joint configurations. Existing approaches use hierarchical planning or latent action representations to enable locomotion, manipulation, and long-horizon loco-manipulation behaviors on real humanoids~\cite{ding2025humanoid, yuan2025being, xue2025leverb, jiang2025wholebodyvla}. Recent open VLA models trained on diverse robot datasets expand the scope of generalized humanoid skills~\cite{gr00tn1_2025}.

Despite these advances, existing humanoid VLA formulations largely abstract away physical interaction forces during planning, leaving them to downstream controllers rather than explicitly specifying the desired interaction regime. Recent studies on tabletop manipulation~\cite{yu2025forcevla, zhang2025ta} incorporate joint-torque or end-effector force feedback into pretrained VLA models and shows improved performance on contact-rich tasks. FWBC-VLA~\cite{zhang2026fwbc} estimates interaction strength from joint-torque residuals and uses this feedback to condition VLA action generation and whole-body compensation on a wheeled-legged platform. Together, these studies demonstrate the value of physical interaction feedback, but such feedback describes the ongoing interaction rather than specifying the force required by the task.
In contrast, our work introduces explicit force commands at the interface between humanoid vision-language planning and whole-body control, enabling desired interaction forces to be predicted from task context and directly regulated during execution.

\section{Method}
We consider contact-rich humanoid manipulation tasks specified by natural language instructions and visual observations, where successful execution requires both geometric motion and appropriate interaction force. Our framework adopts a hierarchical design that separates high-level planning from low-level execution. A VLA policy, instantiated from a pretrained VLA model, predicts geometric motion goals together with contact-force references specifying the desired interaction. An RL-based whole-body controller executes these commands by jointly tracking motion and regulating interaction force. To provide scalable and physically grounded supervision without teleoperation, we use whole-body TO to generate dynamically feasible motion, contact-force, and joint-torque references for controller training. Rollouts of the resulting controllers then provide multimodal training data for VLA fine-tuning.

\subsection{VLA Policy with Explicit Force Commands}
\label{sec:VLA}
Our Opt2VLA generates high-level motion and force commands for whole-body execution.
Given a language instruction $l$, visual observations $I_t$, and histories of robot states $\mathbf{s}_{t-K:t}$ and measured joint torques $\boldsymbol{\tau}_{t-K:t}$, the policy predicts a sequence of geometric motion goals together with temporally aligned contact-force references:
\begin{equation}
\left(
\hat{\mathbf{P}}_{t:t+H}^{\mathrm{ee}},
\hat{\mathbf{F}}^{\mathrm{ref}}_{t:t+H}
\right)
=
\pi_{\mathrm{Opt2VLA}}
\left(
l, I_t, \mathbf{s}_{t-K:t}, \boldsymbol{\tau}_{t-K:t}
\right),
\end{equation}
where $\hat{\mathbf{P}}_{t:t+H}^{\mathrm{ee}}$ denotes the predicted $3$D position waypoints of the left and right hands and feet over horizon $H$, and $\hat{\mathbf{F}}^{\mathrm{ref}}_{t:t+H}$ denotes the corresponding normal-contact-force sequences for the active manipulation contact. 
We instantiate the VLA policy from a pretrained GR00T-N1.7 model~\cite{gr00tn1_2025} and augment its input space with upper-body sensed joint torque history and action space with contact-force references, enabling joint prediction of geometric motion and force commands.

\noindent\textbf{Torque Conditioning.}

We investigate how physical interaction feedback supports the prediction of motion and contact-force references. Inspired by~\cite{zhang2025ta}, we compare incorporating measured joint torques via the native GR00T state encoder against using a dedicated torque encoder that feeds an additional token into the action decoder. We also evaluate whether temporal state and torque histories improves closed-loop execution. For selected variants, we further augment the policy with an auxiliary head that predicts future measured joint torques from the same rollout sequence.
These predictions are supervised using recorded robot torques and are used only as an auxiliary learning objective; the VLA control outputs remain the predicted motion goals and contact-force references. 
We evaluate these choices through closed-loop ablations measuring task completion rate, agreement with language-specified force levels, and force-tracking accuracy.

\noindent\textbf{Data Collection and Language Annotation.}
Training data is collected in simulation by rolling out TO-generated motions using the force-aware controller under different target force conditions. Across trajectories, we vary task configurations such as target force, object pose, and object properties while keeping the corresponding motion reference and force command synchronized. During rollout, we record egocentric RGB observations, robot proprioceptive states, measured joint torques, geometric motion targets, and the associated continuous target-force labels.

Each trajectory is paired with a natural-language instruction describing the task and desired interaction regime. We use the descriptors ``\textit{gently}'', ``\textit{firmly}'', and ``\textit{strongly}'', corresponding to fixed ranges of the continuous target force shared across tasks. For example, prompts include ``Wipe the table \textit{gently}.'', ``Push the box into the shelf \textit{firmly}.'', and ``Pick up the box \textit{strongly}.'' 

\begin{figure*}[!t]
\centering
\includegraphics[width=1\linewidth]{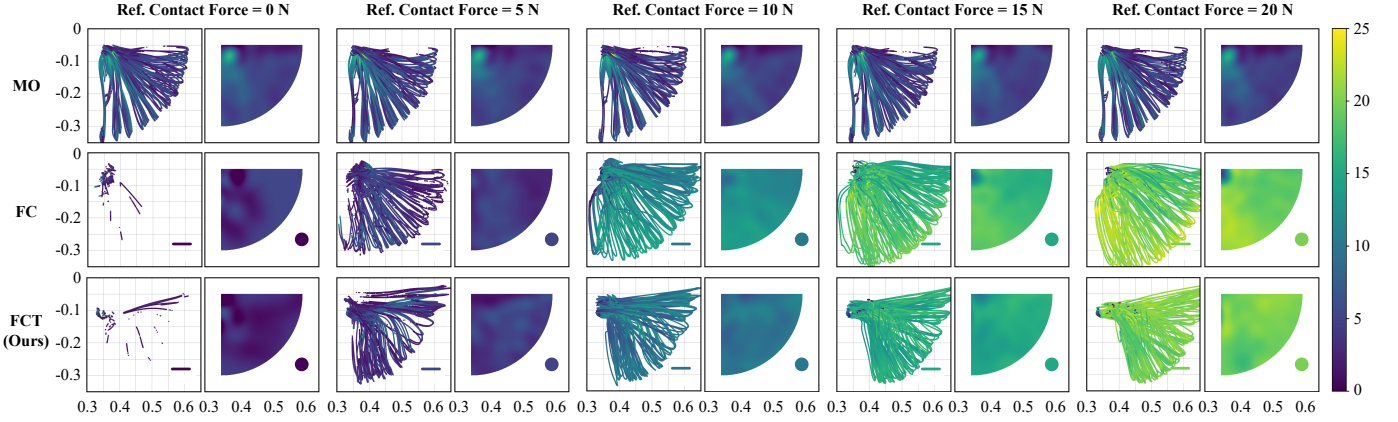}
\vspace{-0.1cm}
\caption{\textbf{Contact force distributions for the surface wiping task under different target forces.}
Each subplot shows scattered measured contact forces (left) across trials at global end-effector $(x,y)$ positions, along with a Gaussian fit (right) summarizing the resulting distribution. Results are aggregated over $10\times3$ trials with varying motion directions and table heights. 
Axes indicate position in meters, and color encodes normal contact force in newtons. FC and FCT produce force levels that vary with the target, whereas MO produces similar forces across target conditions. FCT achieves lower average force-tracking error than FC, as shown in Table~\ref{tab:rl_contactforce}.
}
\label{fig:table_wipe_force_map}
\end{figure*}

\subsection{Force-Aware RL-Based Whole-Body Control}
The motion and contact-force references predicted by the VLA are executed by task-specific FCT controllers.
We formulate each controller as an RL policy that outputs joint-position commands and adopt an asymmetric actor-critic architecture~\cite{pinto2017asymmetric} for sim-to-real transfer.

\noindent\textbf{Policy and Observation.} 
At each timestep $t$, the actor and critic are conditioned on motion and contact-force references:
\begin{equation}
\begin{aligned}
a_t &\sim \pi_\theta
\!\left(a_t \mid o_t^{\mathrm{actor}},
\,
\tilde{\mathbf{p}}_t^{\mathrm{ee}},
\, 
\tilde{\mathbf{F}}_t^{\mathrm{ref}}\right), \\
V_\phi &= 
V_\phi
\!\left(o_t^{\mathrm{critic}},
\, 
\mathbf{p}_t^{\mathrm{ee}},
\, 
\mathbf{F}_t^{\mathrm{ref}}\right),
\end{aligned}
\end{equation}
where ${\mathbf{p}}_t^{\mathrm{ee}}$ denotes the four end-effector motion targets and ${\mathbf{F}}_t^{\mathrm{ref}}$ denotes the target contact force. The tilde \((\tilde{\cdot})\) denotes noisy reference information used to match deployment conditions.

The actor observation $o_t^{\mathrm{actor}}$ consists of deployable proprioceptive signals, including joint states, base velocities, action history, and measured joint torques. The critic observation $o_t^{\mathrm{critic}}$ includes additional privileged simulation states and TO references. Detailed observations are provided in Table~\ref{tab:obs_space}. 

\noindent\textbf{Action Space.} 
The policy outputs joint-position offsets from a default configuration,
$\mathbf{q}^{\mathrm{cmd}}_{t} = 
\mathbf{q}^{\mathrm{dflt}} +
a_t
$
, which are converted to joint torques through a PD controller:
$
\mathbf{u}_t = 
\mathbf{K}_p 
( 
\mathbf{q}^{\mathrm{cmd}}_{t} -
\mathbf{q}_t) - 
\mathbf{K}_d \dot{\mathbf{q}}_t.
$ Here, $\mathbf{q}_t$ denotes the actuated joint positions. The policy learns to adjust these setpoints from motion and contact-force references, proprioception, and measured joint torques, implementing force regulation through the inner PD loop.

\begin{table}[h]
    \centering
    \caption{Observation and state spaces of the force-conditioned controller with torque supervision (FCT).}
    \scriptsize
    \begin{threeparttable}
    \begin{tabular}{lcccc}
        \toprule
        \textbf{Input} & \textbf{Symbol} & \textbf{Dim} & \textbf{Actor} & \textbf{Critic} \\
        \midrule
        Base Lin. Vel.           & $\mathbf{\dot{p}}_{t}^{\rm b}$        & $3$ & \checkmark & \checkmark \\
        Base Ang. Vel.           & $\vg{\omega}^{\rm b}_{t}$         & $3$ & \checkmark & \checkmark \\
        Projected Gravity        & $\mathbf{g}_{t}$       & $3$ & \checkmark & \checkmark \\
        History Motor Joint Pos.   & $\mathbf{q}^{\rm j}_{\rm hist}$     & $200$ & \checkmark & \checkmark \\
        Motor Joint Vel.         & $\mathbf{\dot{q}}^{\rm j}_{t}$              & $20$ & \checkmark & \checkmark \\
        Ref. End-effector Pos.    & $\hat{\mathbf{{p}}}^{\rm ee}_{t}$    & $12$  & \checkmark  & \checkmark \\
        History Action            & $\mathbf{a}_{\rm hist}$             & $200$ & \checkmark & \checkmark \\
        Upper Body Joint Torque   & $\mathbf{u}_{t}$             & $8$ & \checkmark & \checkmark \\
        Ref. Contact Force        & $\hat{\mathbf{{F}}}^{\rm c}_t$             & $2$ & \checkmark & \checkmark \\
 
        \midrule
        Base Translation         & $\mathbf{p^{\rm b}}$        & $3$   &   & \checkmark \\
        Base Orientation         & $\boldsymbol{\theta}^{b}_t$        & $3$   &   & \checkmark \\
        End-effector Pos.        & $\mathbf{p}^{\rm e}$      & $12$  &   & \checkmark \\
        Ref. Base Translation     & $\mathbf{\hat{p}^{\rm b}}$  & $3$   &   & \checkmark \\
        Ref. Base Orientation     & $\hat{\boldsymbol{\theta}}^{b}_t$  & $3$   &   & \checkmark \\
        Ref. Base Lin. Vel.       & $\mathbf{\hat{\dot{p}}^{\rm b}}$  & $3$ &  & \checkmark \\
        Ref. Base Ang. Vel.       & $\hat{\vg{\omega}}^{\rm b}$  & $3$ &  & \checkmark \\
        Ref. Actuator Joint Pos.     & $\mathbf{\hat{q}^{\rm j}}$     & $20$ &  & \checkmark \\
        Ref. Upper Body Joint Torque   & $\hat{\mathbf{u}}_{t}$             & $8$ &  & \checkmark \\
        Contact Force            & $\mathbf{{F}}^{\rm c}_t$             & $2$ &  & \checkmark \\
        PD Gains                 & $\mathbf{K}_p, \mathbf{K}_d$       & $40$  &   & \checkmark\\
        Motor Strength Scale     & $\boldsymbol{\alpha}$  & $20$  &   & \checkmark \\
        
        \bottomrule
\end{tabular}
\label{tab:obs_space}
\begin{tablenotes}[flushleft]
\scriptsize
\item Note that all the force related information including ``Upper Body Joint Torque", ``Ref. Upper Body Joint Torque", ``Contact Force", ``Ref. Contact Force" are not included in the motion-only controller (MO), and ``Ref. Upper Body Joint Torque" is not included in the force-conditioned controller (FC). 
\end{tablenotes}
\end{threeparttable}
\vspace{-0.3cm}
\end{table}

\noindent\textbf{Force-Aware Training.} 
The controller is trained with:
\begin{equation}
\begin{aligned}
r_t = r_t^{\mathrm{motion}} 
      + \lambda_f\, r_t^{\mathrm{force}} 
      + \lambda_\tau\, r_t^{\mathrm{torque}} 
      + r_t^{\mathrm{reg}},
\end{aligned}
\end{equation}
where $r_t^{\mathrm{motion}}$ tracks geometric references, including joint positions, base position and orientation, base linear ad angular velocities, and task-specific end-effector targets. $r_t^{\mathrm{force}}$ tracks the target contact force, $r_t^{\mathrm{torque}}$ tracks TO-derived reference joint torques, and $r_t^{\mathrm{reg}}$ regularizes the motion. Importantly, contact-force references are provided to the actor during both training and deployment, whereas reference torque provides physically grounded guidance only during training.
We follow the domain randomization scheme of~\cite{liu2025opt2skill}, including randomized dynamics, observation noise, control latency, and external perturbations. In addition, we apply Gaussian noise to the motion and force reference observations to improve robustness to prediction errors from the VLA policy.

\noindent\textbf{Deployment.} 
The policy runs at $200$ Hz, with an internal PD loop at $1$ kHz in simulation and $2$ kHz on hardware. At deployment, no TO reference torques are required; the controller tracks the contact-force references predicted by the vision-language planner using onboard observations.

\subsection{Whole-Body Trajectory Optimization for Force Supervision}
Following the TO pipeline in Opt2Skill~\cite{liu2025opt2skill}, we offline whole-body TO to generate supervision for FCT training. Given the robot model, task-specific end-effector motion targets, contact schedule, and prescribed contact wrenches, TO optimizes dynamically feasible whole-body state and joint-torque trajectories subject to floating-base dynamics, contact constraints friction, and joint and actuator limits. We solve the TO using a modified implementation of Crocoddyl and Pinocchio~\cite{mastalli_crocoddyl_2020, carpentier2019pinocchio}, adapted to our contact modeling and constraint requirements. 
The optimized motion trajectory provides geometric references, while the optimized joint torques provide privileged supervision for FCT training;
the prescribed interaction wrench defines the target force command. By varying the task configuration and prescribed force, we generate references across different interaction conditions. The resulting FCT controllers are subsequently rolled out across different task configurations and force conditions to collect multimodal training data for the VLA policy.

\section{Results}
\subsection{Experimental Setup and Tasks}
We evaluate the proposed force-aware VLA framework in simulation and real-world settings on a full-size humanoid robot Digit developed by Agility Robotics. The robot weighs approximately $48$ kg and has $30$ degrees of freedom (DoF), including $20$ actuated joints.

We consider three daily manipulation tasks involving sustained physical interaction under varying contact conditions and object properties:
\begin{enumerate}
    \item \textbf{Surface wiping.} The robot wipes a planar surface using repetitive back-and forth motions while maintaining contact. Appropriate interaction force varies with surface properties and task context.
    \item \textbf{Shelf-box pushing.} The robot pushes a box toward a designated shelf wall and then regulates the contact force against it. Visual observations become partially occluded while physical interaction remains observable.
     \item \textbf{Box pickup with varying weight.} The robot grasps and lifts visually identical boxes with different masses, requiring appropriate grasp forces for reliable and safe manipulation.
\end{enumerate}

Together, these tasks provide different interaction conditions for evaluating force-aware whole-body control and vision-language planning.
Example prompts are provided in Sec.~\ref{sec:VLA}.

\subsection{Force-Aware Whole-Body Control}
We first evaluate Opt2VLA's low-level force regulation to examine how force-conditioned training affects contact-force regulation and whether TO-derived joint-torque supervision provides additional improvements.

\noindent\textbf{Controller variants.} We compare three variants trained with the same RL framework:
\begin{enumerate}
    \item \textbf{Motion-only control (MO).} The controller tracks geometric motion references without target-force inputs or force-tracking objectives.
    \item \textbf{Force-conditioned control (FC).} The controller additionally receives a target interaction force and is trained with a force-tracking objective, without reference joint-torque supervision.
    \item \textbf{Force-conditioned control with torque supervision (FCT, Ours).} In addition to target-force conditioning, FCT uses TO-derived reference joint torques as privileged supervision during training to provide physically grounded guidance for force regulation.
\end{enumerate}

We first evaluate the \textit{surface wiping} task, where similar geometric motions are executed under different target interaction forces. Fig.~\ref{fig:table_wipe_force_map} visualizes the measured contact-force distributions across target force levels and motion directions. Without target-force inputs, MO exhibits large force variability and produces similar distributions across the evaluated force levels.  FC separates the contact force regimes but exhibits larger tracking errors, particularly at lower force levels. In contrast, FCT improves contact-force tracking accuracy and achieves lower average variance, as shown in Table~\ref{tab:rl_contactforce}. 
These results show that explicit force conditioning enables different interaction regimes, while TO-derived torque supervision further improves force regulation accuracy and stability. The torque references are computed jointly with the reference motion under prescribed contact wrenches, subject to whole-body dynamics, contact constraints, and actuator limits. They therefore provide physically consistent joint-level guidance for realizing the desired interaction force.

\begin{figure}[t!]
\centering
\includegraphics[width=1\linewidth]{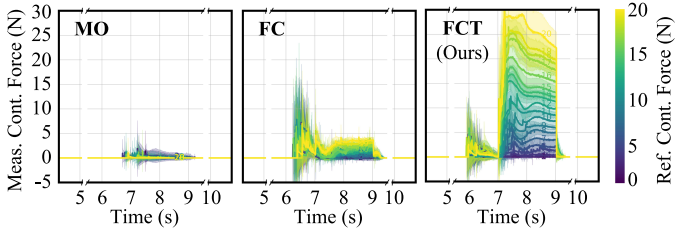}
\vspace{-0.4cm}
\caption{\textbf{Measured push forces for the shelf-box pushing task under different target-force commands}. MO fails to separate force regimes, FC distinguishes force levels but exhibits large tracking error, while FCT achieves more concentrated force distributions and the lowest tracking error. Results at each target force level are aggregated across $10$ target positions to evaluate force tracking over diverse motions, covering $20$ target force levels and $200$ trials in total.
}
\label{fig:RL_shelf_box_force_map}
\end{figure}

The \textit{shelf-box pushing} task exhibits similar but more pronounced differences. The end-effector pushes the box against the designated shelf wall and maintains the specified interaction force. Fig.~\ref{fig:RL_shelf_box_force_map} shows the measured force distributions across different target force levels. MO again produces similar interaction forces across conditions due to the absence of target-force commands, whereas FC separates the target regimes and produces distinct distributions, but exhibits poor force tracking accuracy. FCT achieves the most concentrated distributions and lowest force error ($4.2\pm5.5$ N on average), as shown in Table~\ref{tab:rl_contactforce}.

\begin{figure}[t]
\centering
\includegraphics[width=1\linewidth]{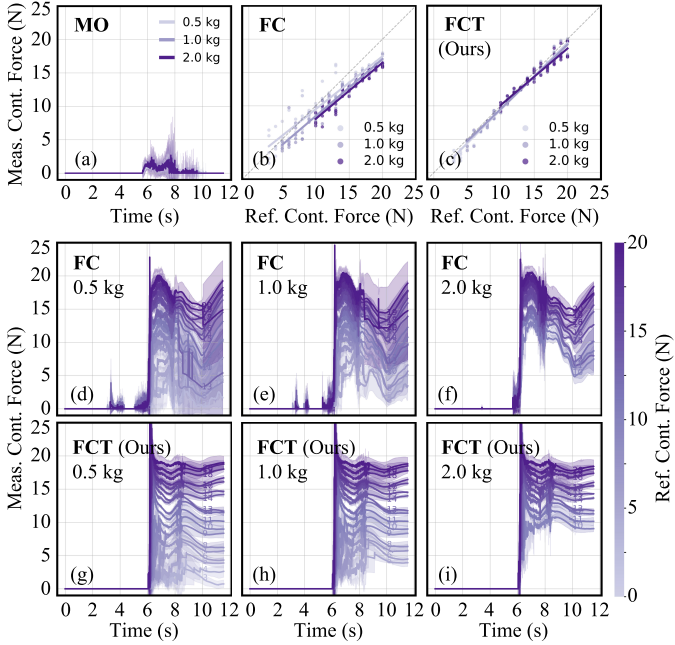}
\vspace{-0.2cm}
\caption{\textbf{Force realization in the box pickup task.} 
(a) Measured grasp force trajectories over time for MO under different object weights. 
(b–c) Realized versus commanded grasp force for FC and FCT across object weights; points denote average lift-and-hold forces with linear fits per weight. 
(d–f) FC grasp force trajectories over time for different object weights and commanded force levels. 
(g–i) Corresponding force trajectories for FCT.
The reference-force ranges are $3$--$20$ N, $5$--$20$ N, and $10$--$20$ N for the $0.5$ kg, $1.0$ kg, and $2.0$ kg boxes, respectively, with $10$ trials per weight–force condition.
}
\vspace{-0.2cm}
\label{fig:RL_box_pickup_force_map}
\end{figure}

Finally, we evaluate the \textit{box pickup} task across varying object weights. All variants receive identical geometric pickup trajectories, while FC and FCT additionally recieve target grasp-force commands spanning $3$--$20$ N.  
As shown in Fig.~\ref{fig:RL_box_pickup_force_map}, MO exhibits similar force behavior across object weights (Fig.~\ref{fig:RL_box_pickup_force_map}(a)) and fails to explicitly regulate the resulting grasp force. In contrast, FC and FCT maintain a clear monotonic relationship between commanded and realized force across different weights (Figs.~\ref{fig:RL_box_pickup_force_map}(b)–(i)). However, FC shows larger tracking errors, particularly for heavier objects. FCT achieves the most consistent force realization, achieving an average grasp-force error of $0.9 \pm 1.0$ N (Table~\ref{tab:rl_contactforce}).

Across these tasks, force-conditioned policies produce contact forces that are consistent with the commanded force values, 
while TO-derived torque supervision further improves force tracking accuracy and consistency across the evaluated contact conditions.

\begin{table}[t]
\setlength{\tabcolsep}{1.1pt}
\renewcommand\arraystretch{1.2}
\caption{\label{tab:rl_contactforce} Force Tracking errors under different contact force references (MAE $\pm$ std in N; Lower Is Better).}
\centering
\scriptsize 
\begin{threeparttable}
\begin{tabular}{ccccccc}
\toprule
\textbf{Task} & \textbf{Ref. Force} & 
\textbf{5 N} & 
\textbf{10 N} & 
\textbf{15 N} & 
\textbf{20 N} & 
\textbf{Avg.}  \\
\midrule
\multirow{2}{*}{\makecell{Surface \\Wiping}}
& FC & 
$3.1 \pm 2.4$ & 
$2.6 \pm 2.5$ & 
$2.5 \pm 1.9$ & 
$1.5 \pm 1.8$ & 
$2.4 \pm 2.9$ \\
& FCT & 
$\mathbf{2.6 \pm 2.6}$ & 
$\mathbf{1.7 \pm 2.1}$ & 
$\mathbf{1.3 \pm 1.5}$ & 
$\mathbf{1.0 \pm 1.2}$ & 
$\mathbf{1.7 \pm 2.1}$ \\
\midrule
\multirow{2}{*}{\makecell{Shelf \\Pushing}}
& FC & 
$4.7 \pm 0.6$ & 
$9.2 \pm 1.1$ & 
$13.7 \pm 1.1$ & 
$17.1 \pm 1.4$ & 
$11.2 \pm 4.8$ \\
& FCT & 
$\mathbf{3.9 \pm 1.2}$ & 
$\mathbf{3.5 \pm 3.4}$ & 
$\mathbf{2.5 \pm 3.5}$ & 
$\mathbf{7.0 \pm 6.7}$ & 
$\mathbf{4.2 \pm 5.5}$ \\
\midrule
\multirow{2}{*}{\makecell{Box \\Pickup}}
& FC & 
$1.0 \pm 1.2$ & 
$1.6 \pm 0.9$ & 
$2.5 \pm 0.6$ & 
$3.5 \pm 0.6$ & 
$2.3 \pm 1.4$ \\
& FCT & 
$\mathbf{0.7 \pm 0.5}$ & 
$\mathbf{0.5 \pm 0.5}$ & 
$\mathbf{0.6 \pm 0.7}$ & 
$\mathbf{1.6 \pm 1.1}$ & 
$\mathbf{0.9 \pm 1.0}$ \\
\midrule
\end{tabular}
\end{threeparttable}
\end{table}

\subsection{Vision-Language Planning with Explicit Force Commands}
\label{sec:vla_main_eval}

We evaluate closed-loop Opt2VLA execution using nine multi-task VLA configurations that predict motion and contact-force references and share the same task-specific FCT controllers. The configurations differ in torque encoding, observation history, auxiliary future-torque supervision, and backbone fine-tuning. Each variant is evaluated on ten reference-defined scenes per task under \textit{gentle}, \textit{firm}, and \textit{strong} prompts, yielding $90$ episodes.
Scene appearances use observed training-appearance families. Scenes, prompts, and inference settings are shared across variants.

\begin{table}[!t]
\centering
\caption{Closed-loop Opt2VLA ablation: joint success rates (\%). Task columns average the three force prompts.}
\label{tab:vla_nominal_results}
\scriptsize
\setlength{\tabcolsep}{1.25pt}
\renewcommand{\arraystretch}{1.15}
\begin{threeparttable}
\begin{tabular}{@{}lclccrrrr@{}}
\toprule
Variant & $H_s$ & Torque & Aux. & Train
& \makecell{Box\\pickup} & \makecell{Shelf\\pushing.} & \makecell{Surface\\wiping} & Overall \\
\midrule
Baseline (C0) & 1 & -- & -- & T-H & 33.3 & 40.0 & 46.7 & 40.0 \\
+ future-$\tau$ (C1) & 1 & -- & $\checkmark$ & T-H & 33.3 & 60.0 & 63.3 & 52.2 \\
State torque (C2) & 1 & State, 1 & -- & T-H & 50.0 & 46.7 & 50.0 & 48.9 \\
+ future-$\tau$ (C3) & 1 & State, 1 & $\checkmark$ & T-H & 56.7 & 50.0 & 66.7 & 57.8 \\
State history (C4) & 10 & State, 10 & -- & T-H & 6.7 & 43.3 & 33.3 & 27.8 \\
+ future-$\tau$ (C5) & 10 & State, 10 & $\checkmark$ & T-H & 13.3 & 66.7 & 33.3 & 37.8 \\
Token history (C6) & 10 & Token, 10 & -- & T-H & 53.3 & \underline{70.0} & \underline{100.0} & 74.4 \\
\textbf{+ future-$\tau$ (C7, Ours)} & 10 & Token, 10 & $\checkmark$ & T-H & \underline{76.7} & \underline{70.0} & \underline{100.0} & \underline{82.2} \\
Full FT (C8) & 10 & Token, 10 & $\checkmark$ & Full & \underline{76.7} & 50.0 & \underline{100.0} & 75.6 \\
\bottomrule
\end{tabular}
\begin{tablenotes}[flushleft]
\scriptsize
\item $n=30$ per task and $90$ overall. Underlines mark the best rates, including ties. 

C7 is the default Opt2VLA planner configuration; all variants use the same task-specific FCT controllers.
\item $H_s$: state-history length. Torque entries give the encoder and history length: State, concatenation with state; Token, a dedicated torque encoder. Ten-sample histories use $200$-ms spacing. Aux.: future measured-torque prediction.
\item T-H: frozen VLM with connectors, action model, and VLM normalization trained; Full also trains the VLM. 
\item Each row represents one VLA model evaluated across all three tasks.
\end{tablenotes}
\end{threeparttable}
\vspace{-0.2cm}
\end{table}

\noindent\textbf{Evaluation Criteria.}
We assess task completion, prompt-band agreement, and mean-force agreement.
Pickup requires the lowest point of the box to clear the tabletop by at least $5$~cm; shelf-box pushing requires reaching the designated right-side wall; and wiping requires a continuous tabletop-contact stroke of at least $5$~cm within $30^\circ$ of the vertical wiping axis.
For each episode, both force criteria use samples pooled across all command-active windows, without conditioning on measured contact. Prompt-band agreement requires each evaluated hand's mean commanded force to fall within the training-defined range associated with the instruction, allowing a $2$~N tolerance at each boundary. The resulting acceptance bands are $[0,9]$~N, $(5,16]$~N, and $(12,22]$~N for \textit{gentle}, \textit{firm}, and \textit{strong}, respectively.
Mean-force agreement requires the pooled commanded and measured means to differ by at most $5$~N per hand. Joint success requires task completion in a command-associated attempt and both episode-level force criteria; both hands must pass for pickup. Episodes without a command-active window fail, and SR includes all evaluated episodes. Appendix~\ref{sec:vla_eval_details} provides the window and reset definitions.

\begin{figure}[t]
\centering
\includegraphics[width=1\linewidth]{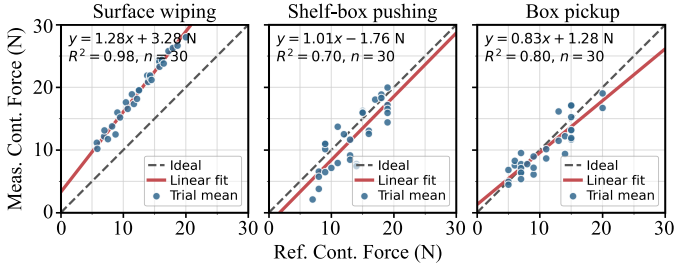}
\vspace{-0.3cm}
\caption{\textbf{Hardware force tracking of Opt2VLA's FCT controller.} 
(a) surface wiping, (b) shelf-box pushing, and (c) box pickup. Markers show per-trial mean, baseline-corrected force during steady contact; dashed lines indicate ideal tracking and solid lines show least-squares fits.}
\label{fig:RL_hardware_force_tracking}
\end{figure}

\begin{figure}[t]
\centering
\includegraphics[width=1\linewidth]{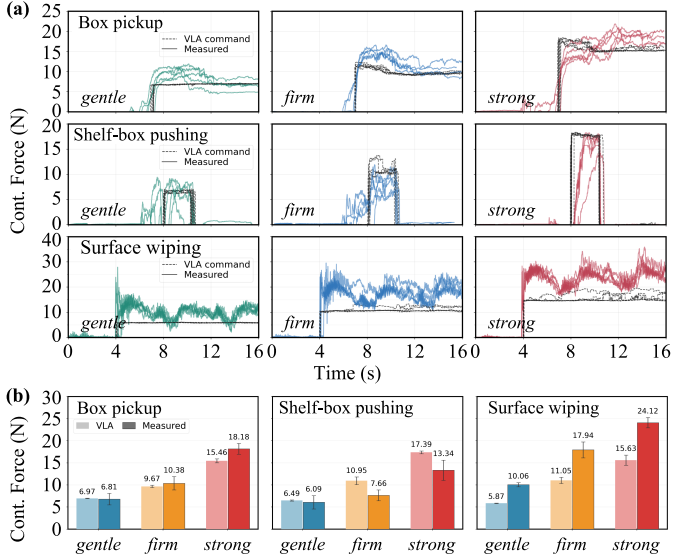}
\vspace{-0.3cm}
\caption{\textbf{End-to-end Opt2VLA force modulation on hardware.}
(a) VLA commands (dashed) and measured forces (solid).
(b) Mean commanded and measured forces over the task-specific evaluation phase. Results include five trials per task and prompt.}
\label{fig:vla_hardware_force_tracking}
\vspace{-0.3cm}
\end{figure}

\noindent\textbf{Language-Conditioned Motion--Force Execution.}
Table~\ref{tab:vla_nominal_results} summarizes joint success.
Completing the task does not necessarily satisfy the requested force level: the kinematic-input variant (C0) passes the task gate in $81.1\%$ of episodes, decreasing to $53.3\%$ when prompt-band agreement is also required and $40.0\%$ under joint scoring.
Opt2VLA (C7) achieves $90.0\%$ for task completion with prompt-band agreement and the highest overall joint SR of $82.2\%$, indicating stronger agreement between task execution, language-specified force levels, and realized interaction.

\noindent\textbf{Torque-Aware Planning.}
The benefit of interaction feedback depends on its representation and supervision. With the same observation histories and no auxiliary loss, a dedicated torque encoder (C6) achieves a joint SR of $74.4\%$, compared with $27.8\%$ for native state concatenation (C4). Adding future-torque supervision (C7) increases overall joint SR to $82.2\%$, with pickup improving from $53.3\%$ to $76.7\%$ and the other two task rates unchanged.

The improvement is task-dependent: C6 and C7 share the highest \textit{shelf-box pushing} SR of $70.0\%$. Full-backbone fine-tuning (C8) matches C7 on \textit{box pickup} and \textit{surface wiping}, but achieves $50.0\%$ on \textit{shelf-box pushing}, resulting in an overall joint SR of $75.6\%$. Overall, Opt2VLA (C7) achieves the highest joint SR across the evaluated tasks using a dedicated torque encoder, observation histories, and auxiliary future-torque supervision with a frozen vision-language backbone.

\subsection{Hardware Evaluation}

\noindent\textbf{RL force tracking.}
We evaluate force tracking by Opt2VLA's FCT controller over $30$ hardware trials per task, as shown in Fig.~\ref{fig:RL_hardware_force_tracking}.
For surface wiping, the normal-force error is computed over the detected sustained-contact phase. With a spanning reference forces of $5.0$--$20.0$~N, the controller achieves a mean absolute error (MAE) of $7.0$~N ($95\%$ bootstrap CI: $6.5$--$7.4$~N). The measured force remained correlated with the reference ($R^2=0.98$), showing consistent modulation across commanded force levels.

For shelf-box pushing, force is evaluated over the sustained-contact phase. the controller achieves an MAE of $3.9$~N ($95$\% bootstrap CI: $3.6$--$4.3$ N). The realized force varied approximately linearly with the reference with $R^2=0.70$, indicating that the controller reproduced the overall trend of the commanded force across force levels.
For box pickup, we evaluate the steady-state force tracking over the final $2$~s of the holding phase. The controller achieves an MAE of $1.7$ N ($95$\% bootstrap CI: $1.4$--$2.1$ N). The measured force exhibits a strong relationship with the reference force, with a calibration slope of $0.83$ and $R^2=0.80$.

Overall, FCT preserves the commanded force ordering and consistent force modulation after sim-to-real transfer, although the absolute tracking accuracy varies across tasks.

\noindent\textbf{End-to-End Opt2VLA execution.}
We evaluate the complete Opt2VLA system over $45$ hardware trials, using the same VLA checkpoint across all three tasks, with five trials per task under \textit{gentle}, \textit{firm}, and \textit{strong}
prompts (Fig.~\ref{fig:vla_hardware_force_tracking}).
The mean VLA-issued force commands fall within the intended force regimes across all three tasks.
Surface wiping exhibits a larger command-to-measurement gap, consistent with the tracking bias observed in the RL-only hardware evaluation. This suggests that low-level force-tracking error contributes to the gap even when the VLA selects the intended force level. Overall, these results demonstrate language-conditioned force modulation during closed-loop hardware execution.

\section{Conclusion} 
\label{sec:conclusion}
We present Opt2VLA, a force-aware VLA framework that introduces explicit force commands as the interface between vision-language planning and humanoid whole-body control. The policy jointly predicts geometric motion goals and contact-force references, which are tracked by an RL-based controller trained to regulate motion and interaction force. Whole-body TO which prescribed contact wrenches provides dynamically feasible motion and joint-torque references for physically grounded controller training and scalable multimodal data generation. Experiments on three humanoid whole-body manipulation tasks show that explicit force conditioning enables distinct interaction regimes beyond geometric tracking alone, while TO-derived torque supervision further improves force-tracking accuracy and consistency. Closed-loop VLA evaluation further shows that by adding torque history via a separate encoder and adding an auxiliary future torque loss, we can achieve much better force command generation and force command tracking.
Hardware experiments demonstrate the sim-to-real transfer of the Opt2VLA framework to a fill-size humanoid robot.


\bibliographystyle{IEEEtran}
\bibliography{bibliography}

\clearpage
\onecolumn
\section{Appendix}
\subsection{Reward Design and Domain Randomization}

\begin{table}[h]
\caption{Reward Components and Weights.}
\centering
\small
\label{tab:reward}
\begin{threeparttable}
\begin{tabular}{l l l c}
\midrule
\textbf{Category} 
& \textbf{Term} 
& \textbf{Expression} 
& \textbf{Weight} \\

\midrule
\multirow{8}{*}{\makecell{Task \\ Reward}}
&Joint Pos. 
& $\exp(-5 \|\mathbf{{\hat{q}}}^{\rm j}_t - \mathbf{{{q}}}^{\rm j}_t\|_2^2)$                 
& $3$ \\
& Base Pos. 
& $\exp(-20 \|\mathbf{\hat{p}}^{\rm b}_t - \mathbf{{p}}^{\rm b}_t\|_2^2)$                 
& $3$ \\
& Base Ori. 
& $\exp(-50 \|\hat{\boldsymbol{\theta}}^{\rm b}_t - {\boldsymbol{\theta}}^{\rm b}_t\|_2^2)$       
& $3$ \\
& Base Lin. Vel. 
& $\exp(-2 \| \mathbf{\hat{\dot{p}}}_t^{\rm b} - \mathbf{{\dot{p}}}^{\rm b}_t\|_2^2)$             
& $3$ \\
& Base Ang. Vel. 
& $\exp(-0.5 \|\hat{\boldsymbol{\omega}}^{\rm b}_t - {\boldsymbol{\omega}}^{\rm b}_t\|_2^2)$ 
& $3$ \\
& End-effector Pos. 
& $\exp(-20 \|\hat{\mathbf{{p}}}^{\rm ee}_t - {\mathbf{{p}}}^{\rm ee}_t\|_2^2)$         
& $3$ \\
& Joint Torque 
& $\exp(-0.01 \|\hat{\mathbf{{u}}}_t - {\mathbf{{u}}}_t\|_2^2)$         
& $2$ \\
& Contact Force 
& $\exp(-0.05 \|\hat{\mathbf{{F}}}^{\rm c}_t - {\mathbf{{F}}}^{\rm c}_t\|_1)$         
& $2$ \\
\midrule
\multirow{3}{*}{\makecell{Penalty\\ Cost}} 
& Action Rate 
& $\|\mathbf{a}_t - 2\mathbf{a}_{t-1} + \mathbf{a}_{t-2}\|_2^2$            
& $-3$ \\
& Torques 
& $\|\mathbf{u}_t / \mathbf{u}_{\text{limit}}\|_2^2$                                        
& $-0.3$ \\
& Joint Acc. 
& $\|\mathbf{{\ddot{q}}}^{\rm j}_t\|_2^2$                                        
& $-10^{-5}$ \\
\midrule
\end{tabular}
\begin{tablenotes}[flushleft]
\small
\item Note that the ``Joint Torque" and ``Contact Force" rewards are not included in the MO controller, and ``Joint Torque" reward is not included in the FC controller.
\end{tablenotes}
\end{threeparttable}
\end{table}

\begin{table}[h]
\caption{Domain Randomization Parameters.}
\centering
\small
\label{tab:domain_randomization}
\begin{tabular}{l l l l}
\toprule
\textbf{Category} & \textbf{Parameter} & \textbf{Type} & \textbf{Range / Std} \\
\midrule
\multirow{5}{*}{Obs.} 
& Ref. End-effector Pos. (m) 
& Additive (Gauss) 
& $\sigma=0.02$ \\
& Ref. Contact Force (N) 
& Additive (Gauss) 
& $\sigma=2.0$ \\
& Joint Pos. (rad) 
& Additive (Gauss) 
& $\sigma=0.0875$ \\
& Joint Vel. (rad/s)
& Additive (Gauss) 
& $\sigma=0.075$ \\
& Base Lin. Vel. (m/s)
& Additive (Gauss) 
& $\sigma=0.075$ \\
& Base Ang. Vel. (rad/s)
& Additive (Gauss) 
& $\sigma=0.075$ \\
& Gravity Proj. (m/s$^2$) 
& Additive (Gauss) 
& $\sigma=0.0375$ \\
\midrule
Delays 
& Action Delay (s)
& Uniform 
& $[0.0, 0.02]$ \\
\midrule
\multirow{2}{*}{Motor} 
& Motor Strength 
& Scaling (Uniform) 
& $[0.95, 1.05]$ \\
& Kp/Kd Factor 
& Scaling (Uniform) 
& $[0.9, 1.1]$ \\
\midrule
\multirow{3}{*}{Reset} 
& Joint pos. init. (rad) & Additive (Gauss) & $\sigma=0.03$ \\
& Joint vel. init. (rad/s) & Additive (Gauss) & $\sigma=0.06$ \\
& Root vel. init. (m/s) & Additive (Gauss) & $\sigma=0.05$ \\
\midrule
\multirow{2}{*}{Env.} 
& Gravity (m/s$^2$) 
& Scaling (Uniform) 
& $[0.9, 1.1]$ \\
& Wipe: Table height offset (m) 
& Additive (Uniform)
& $[-0.01, 0.01]$ \\
\bottomrule
\end{tabular}
\end{table}

\subsection{Whole-Body Trajectory Optimization for Force Supervision}
We use whole-body TO offline to generate physically grounded supervision for force-aware planning and control. Given task-specific motion targets and prescribed contact wrenches, TO generates dynamically feasible trajectories that satisfy whole-body dynamics and contact constraints. This provides geometric motion and joint-torque references that are physically consistent with the desired interaction forces, which are difficult to obtain from kinematic planning or human demonstrations. 

We consider a standard floating-base humanoid model with generalized coordinates $\mathbf{q}$ and velocities $\mathbf{v}$, governed by the equations of motion:
\begin{equation} 
\label{eq:fbEoM}
\begin{aligned}
\mathbf{M}(\mathbf{q}) 
\mathbf{\dot{v}} + 
\mathbf{C}
(\mathbf{q, v}) = 
{\mathbf{B}} 
\v u + 
\mathbf{J_{\rm c}^{\top}} 
\mathbf{F_{\rm c}},
\end{aligned}
\end{equation}
where $\mathbf{M(q)}$ is the joint-space mass matrix, $\mathbf{C}
(\mathbf{q, v})$ collects Coriolis, centrifugal, and gravitational terms, ${\mathbf{B}}$ is the actuation matrix, $\v u$ denotes joint torques, $\mathbf{J_{\rm c}(q)}$ is the contact Jacobian, and $\mathbf{F_{\rm c}}$ denotes contact wrench forces.

Over a finite horizon, TO optimizes whole-body motion and joint torques under prescribed interaction wrenches:
\begin{subequations} \label{eq:whole_to}
\begin{align} 
\label{eq:cost}
& 
\underset{
\mathbf{x},
\mathbf{u}
}
{\min} 
& &
\sum_{k=0}^{N - 1} 
\Bigl( 
\| 
\mathbf{y}[k] - 
\hat{\mathbf{y}}[k] \|^2 _Q + 
\| 
\mathbf{u}[k] \|^2_{R} 
\Big) + \\
&
&& 
\| 
\mathbf{y}[N] - 
\hat{\mathbf{y}}[N] \|^2 _{Q_{\rm f}} \nonumber \\
&
\text{subject to} \nonumber \\
\label{eq:hybrid_dynamics}
&
(\textit{\rm Dynamics})         
&& 
\begin{cases} 
\v M(\v q) 
\v {\ddot q} + 
\v C(\v q, \v {\dot q})  = 
\v B 
\v u + 
\mathbf{J_{\rm c}^{\top}} 
\mathbf{F_{\rm c}}\\
\v M(\v q)\v{\dot{q}}^+ - 
\v M(\v q)\v{\dot{q}}^- = 
\mathbf{J_{\rm c}^{\top}} 
\vg 
\Lambda
\end{cases}\\
\label{eq:contact}
&
(\textit{\rm Contact}) 
&& 
\begin{cases}
\v J_{\rm c} 
\v{\ddot{q}}+
\mathbf{\dot{J}_{\rm c}}
\v{\dot{q}} = 
0\\
\textcolor{black}{
\v J_{\rm c} 
\v{\dot{q}}^+ = 
0
}
\end{cases}\\
\label{eq:joint_torque_limits}
&
(\textit{\rm Limits}) 
&& 
\v q^{\rm j} 
\in \mathcal{J}, 
\v u \in \mathcal{T}\\
&
\text{(Friction)} 
&&
\mathbf{F}_{\rm c} 
\in \mathcal{F}(\mu, \mathbf{q}). 
\hspace{1cm} 
\label{eq:friction_cone}
\end{align}
\end{subequations}

Here $\mathbf{y} = \Phi(\mathbf{q})$ denotes task-space quantities such as end-effector pose. we solve the optimization using a differential dynamic programming (DDP)-based method implemented in Crocoddyl and Pinocchio~\cite{mastalli_crocoddyl_2020, carpentier2019pinocchio}.

\subsection{Hardware Force-Tracking Evaluation}

\noindent\textbf{Force-sensing setup and preprocessing.}
External force sensors were used only to evaluate hardware force tracking and were not provided to the control policy. For surface wiping, the vertical normal component $F_z$ was obtained from a Bertec Force Plate ($40$ cm $\times60$ cm). For shelf-box pushing and box pickup, force was measured using a SensX 160 force sensor ($65$ mm $\times25$ mm). All signals were baseline-corrected. Filtering was used only to identify contact-phase boundaries; all force-tracking metrics were computed from the unfiltered, baseline-corrected measurements.


\begin{figure*}[t]
    \centering
        \includegraphics[width=0.6\linewidth]
        {{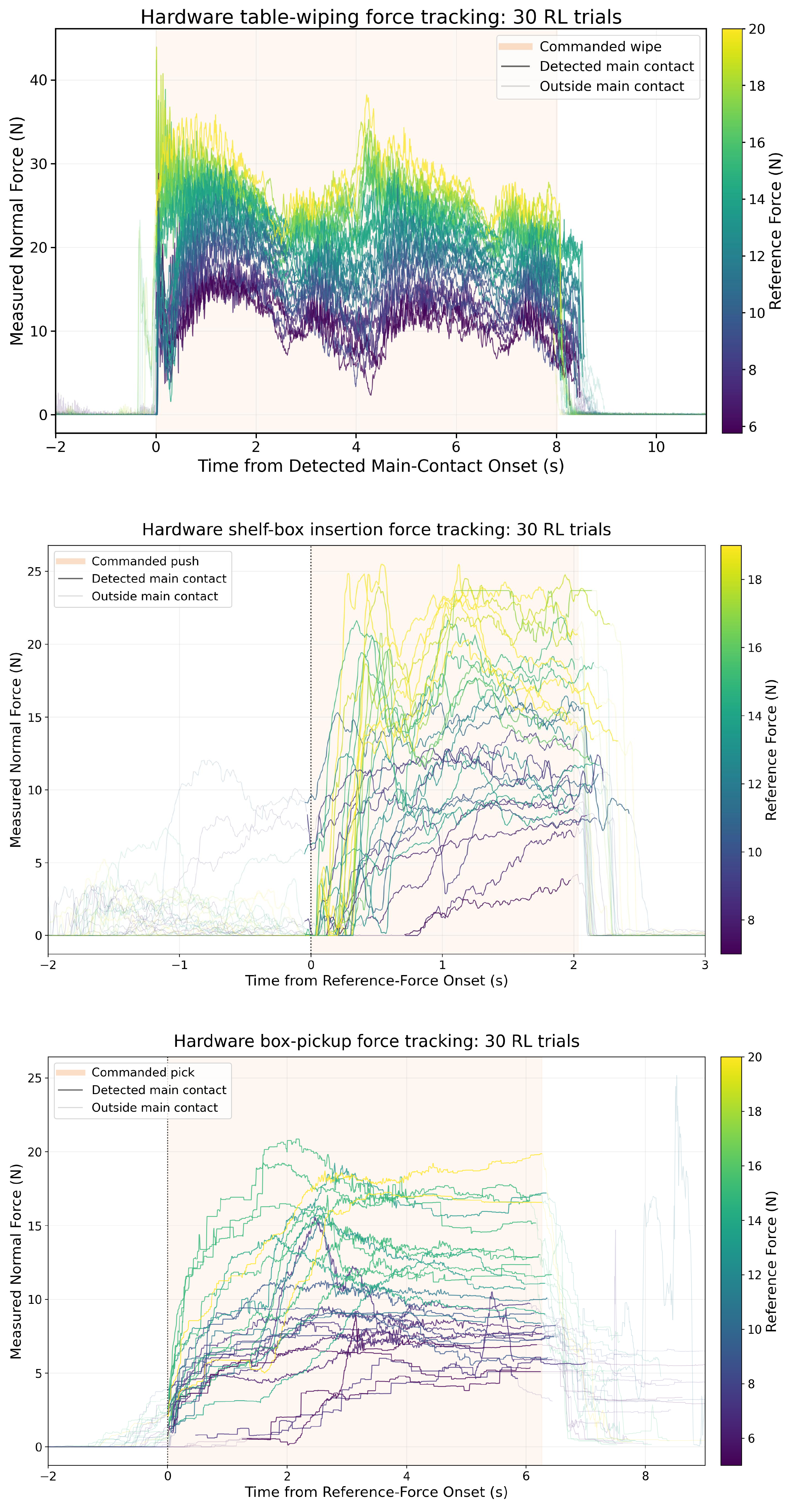}}
        \caption{\textbf{Hardware force trajectories across all trials.}
        Baseline-corrected normal-force trajectories are shown for $30$ trials per task. Trajectory color indicates the reference force. Opaque segments denote the sustained-contact interval used for analysis, faded segments show measurements outside main contact, and orange shading indicates the commanded-force phase. Differences in trace smoothness reflect the sensing modalities used across tasks.}
    \label{fig:supp_all_force_trials}
\end{figure*}

\begin{figure*}[t]
    \centering
        \includegraphics[width=0.8\linewidth]
        {{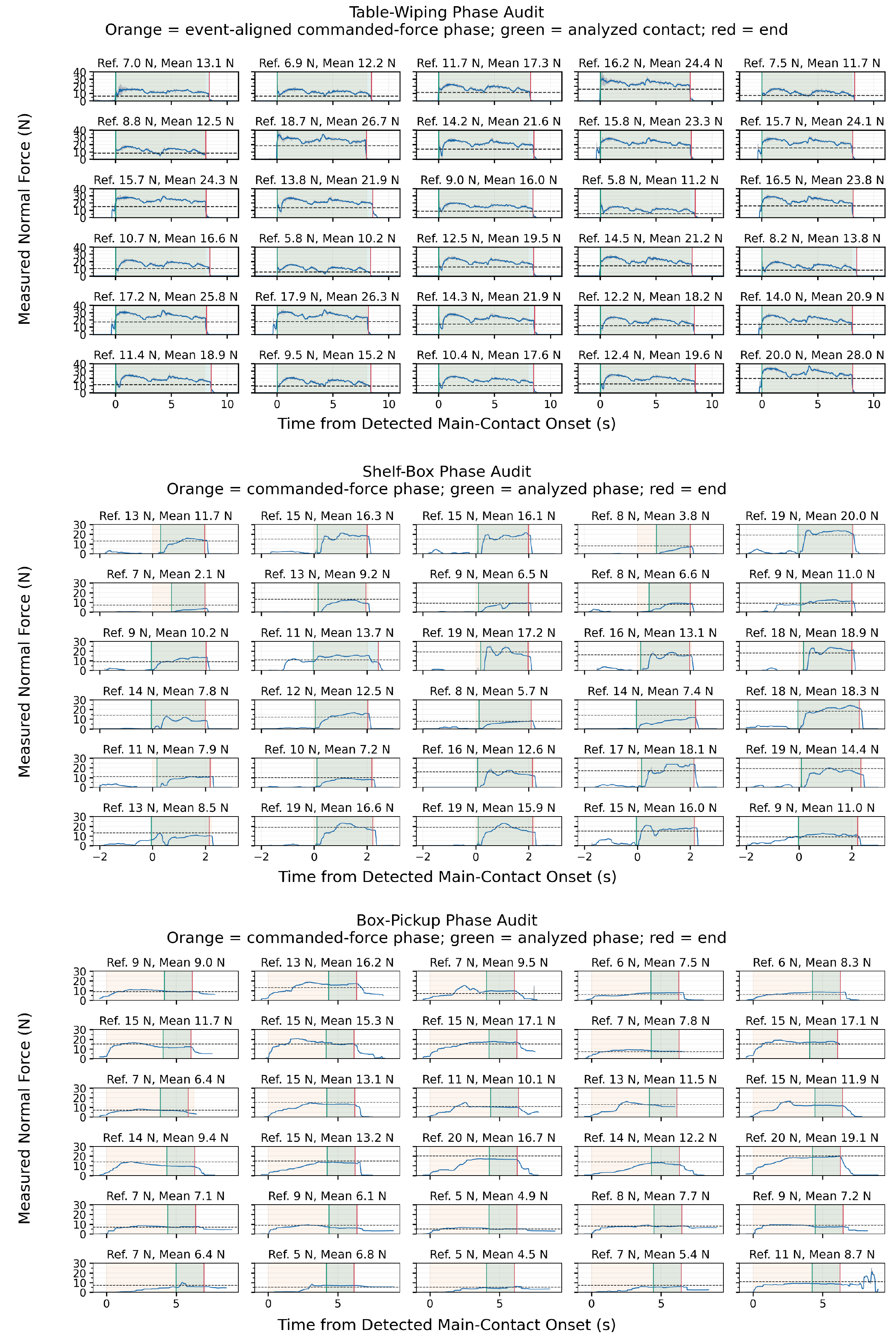}}
        \caption{\textbf{Contact-phase detection for the hardware trials.}
        Each subplot presents one trial. Gray curves show the unfiltered, baseline-corrected normal force; blue curves show the filtered signal used only for phase-boundary detection, and dashed black lines indicate the reference force. Orange and green shading shows the commanded-force phase and the retained analysis interval, respectively. The green and red vertical lines indicate the beginning and end of the analyzed interval.} 
    \label{fig:supp_all_force_trials_phase}
\end{figure*}

\clearpage
\begingroup
\makeatletter
\setlength{\@fptop}{0pt}
\setlength{\@fpsep}{12pt}
\setlength{\@fpbot}{0pt plus 1fil}
\makeatother
\subsection{Closed-Loop VLA Simulation Protocol}
\label{sec:vla_eval_details}

\noindent\textbf{Evaluation design.}
We evaluate each variant using a single $40{,}000$-step checkpoint shared across all three tasks, with a fixed task-specific FCT controller. Each task uses ten reference-defined scenes and three force-level prompts: $30$ episodes per task and $90$ episodes per variant. Scene initialization, prompts, low-level controllers, and inference settings are shared across variants. Simulator and inference seeds are fixed to $123$ and $0$, respectively; the three prompts for a given scene use the same initial reset.

\noindent\textbf{Scene and robot initialization.}
Reference trajectories specify support and object placement, nominal robot base pose, actuated-joint state, and initial velocities; the XML supplies remaining scene geometry and robot coordinates. Shelf scenes contain a fixed cabinet and a movable box.

For pickup and wiping, robot resets add zero-mean Gaussian noise with standard deviation $0.03$~rad to actuated arm and leg joint positions. Shelf-box pushing retains the same arm-position perturbations but initializes leg positions from the environment's prescribed posture. All three tasks sample initial base linear, base angular, and actuated-joint velocities from zero-mean Gaussian distributions with standard deviations $0.05$~m/s, $0.05$~rad/s, and $0.06$~rad/s, respectively; shelf leg velocities remain perturbed. Base position and orientation are not independently randomized, and other domain randomization is disabled.

\noindent\textbf{Language and appearance conditions.}
The prompt templates are \emph{Pick up the box \{gently/firmly/strongly\}.}, \emph{Push the box into the shelf \{gently/firmly/strongly\}.}, and \emph{Wipe the table with \{gentle/firm/strong\} pressure.} One descriptor is substituted for each episode without additional direction qualifiers.

Table~\ref{tab:vla_appearance_conditions} lists the appearance configurations. Each appearance is assigned to two reference scenes and reused across all three force prompts and VLA variants. Floors and physical geometry remain unchanged.

\begin{table}[!htbp]
\centering
\caption{Evaluation appearance configurations. Pickup and shelf entries show support/box pairs; wiping entries describe the table appearance. Each appearance is assigned to two reference scenes.}
\label{tab:vla_appearance_conditions}
\small
\begin{tabularx}{\linewidth}{@{}lX@{}}
\toprule
Task & Appearance configurations \\
\midrule
Box pickup & White/yellow; yellow/white; blue/white; tan/blue; tan/yellow \\
\midrule
Shelf pushing & White/yellow; yellow/blue; tan/blue; blue/tan; gray/yellow \\
\midrule
Table wiping & White; gray-metal (textured); wood (textured); blue; yellow \\
\bottomrule
\end{tabularx}
\end{table}

\noindent\textbf{Execution and observations.}
MuJoCo/MJX runs batches of four environments with native $320\times240$ chest-camera rendering. Physics integration, low-level control, issued VLA commands, and VLA inference run at $1{,}000$, $200$, $50$, and $5$~Hz, respectively. Each variant uses its trained observation schema; ten-sample state and torque histories use $200$-ms spacing, with unavailable initial history filled by repeating the first observation. Future-torque targets are used only for auxiliary training and are not inference inputs.

Before VLA control begins, the task-specific FCT controller follows reference commands for a $1.8$-s initialization period, which is excluded from scoring. Episode horizons are $20$~s for pickup, $15$~s for shelf-box pushing, and the reference endpoint ($15.3$--$15.6$~s) for wiping, subject to early termination.

\noindent\textbf{Task-success criteria.}
Pickup succeeds when the lowest point of the oriented box collision geometry clears the tabletop by at least $5$~cm, without a tilt constraint. Shelf-box pushing succeeds when the box reaches the designated right-wall, defined by wall overlap in world $X/Z$ and a signed wall gap in $[-5,+0.01]$~mm. Wiping succeeds when the robot produces one continuous tabletop-contact stroke of at least $5$~cm within $30^\circ$ of the reset-heading forward--backward axis; either direction qualifies. Contact loss separates strokes, and a $1$-cm reversal hysteresis separates successive wiping motions. No task requires goal-state dwell.

\noindent\textbf{Force intervals and gates.}
Both hands are evaluated for pickup, while only the right hand is evaluated for shelf-box pushing and wiping. A command-active window starts when the maximum issued force over the evaluated hands exceeds $0.5$~N continuously for $0.1$~s, and ends when it remains at or below $0.5$~N for $0.2$~s. Detected boundaries are backdated to the beginning of the qualifying sequence; shorter interruptions remain included. Windows still active at the end of VLA control are retained up to that endpoint.

For each hand, commanded and measured force means pool all controller-rate samples in the union of the episode's command-active windows. Longer windows contribute proportionally more samples; intervening inactive gaps are excluded. Measured zero-force samples remain included, and measured contact does not determine the windows. Both pickup hands use the same union but are scored separately.

Prompt-band agreement requires each hand's pooled commanded mean to lie in $[0,9]$~N, $(5,16]$~N, or $(12,22]$~N for \emph{gentle}, \emph{firm}, or \emph{strong}, respectively, allowing $2$~N beyond each nominal training-range boundary and truncating at zero. Mean-force agreement requires the absolute difference between pooled measured and commanded means to be at most $5$~N per hand. Reference-trajectory forces are not scoring targets. Both pickup hands must pass independently; an episode without a command-active window fails both force criteria.

\noindent\textbf{Repeated attempts and joint success.}
Each command-active window defines a task-attempt interval extending from its start until the next command window begins or active control ends. The task gate passes if any such interval meets the task criterion. Both force gates instead use the union of all command-active windows, including unsuccessful attempts, with no selection of a passing window. Joint success requires the task gate and both pooled force gates. The task-stage rates below use this command-associated task gate, without force requirements.

\noindent\textbf{Aggregation and tracking diagnostics.}
Prompt-, task-, and overall-level SRs use $10$, $30$, and $90$ episodes per variant, respectively, with equal weighting across task--prompt combinations. Early-terminated episodes remain in the denominator and are scored on the available active-control samples. Table~\ref{tab:vla_detailed_results} reports prompt-level joint SR and cumulative task, task-plus-intent, and joint rates.

\begin{table}[!htbp]
\centering
\caption{Detailed closed-loop VLA results (\%). Task blocks report joint success by force prompt and task average. Overall columns report cumulative task-gate, task-plus-prompt-band, and joint success. Both force gates pool all command-active windows.}
\label{tab:vla_detailed_results}
\small
\setlength{\tabcolsep}{3pt}
\renewcommand{\arraystretch}{1.12}
\begin{tabular}{@{}l*{15}{r}@{}}
\toprule
& \multicolumn{4}{c}{Box pickup} & \multicolumn{4}{c}{Shelf pushing}
& \multicolumn{4}{c}{Surface wiping} & \multicolumn{3}{c}{Overall} \\
\cmidrule(lr){2-5}\cmidrule(lr){6-9}\cmidrule(lr){10-13}\cmidrule(lr){14-16}
Variant & G & F & S & Avg. & G & F & S & Avg. & G & F & S & Avg.
& Task & + band & Joint \\
\midrule
C0 & \underline{100.0} & 0.0 & 0.0 & 33.3 & 40.0 & 30.0 & 50.0 & 40.0 & \underline{100.0} & 40.0 & 0.0 & 46.7 & 81.1 & 53.3 & 40.0 \\
C1 & 80.0 & 20.0 & 0.0 & 33.3 & 60.0 & \underline{80.0} & 40.0 & 60.0 & 90.0 & 90.0 & 10.0 & 63.3 & 78.9 & 57.8 & 52.2 \\
C2 & 70.0 & 80.0 & 0.0 & 50.0 & \underline{70.0} & 40.0 & 30.0 & 46.7 & 80.0 & 70.0 & 0.0 & 50.0 & 81.1 & 53.3 & 48.9 \\
C3 & 90.0 & 80.0 & 0.0 & 56.7 & 30.0 & 50.0 & 70.0 & 50.0 & \underline{100.0} & \underline{100.0} & 0.0 & 66.7 & 75.6 & 63.3 & 57.8 \\
C4 & 20.0 & 0.0 & 0.0 & 6.7 & 30.0 & 20.0 & 80.0 & 43.3 & \underline{100.0} & 0.0 & 0.0 & 33.3 & 64.4 & 37.8 & 27.8 \\
C5 & 40.0 & 0.0 & 0.0 & 13.3 & 30.0 & \underline{80.0} & \underline{90.0} & 66.7 & \underline{100.0} & 0.0 & 0.0 & 33.3 & 70.0 & 47.8 & 37.8 \\
C6 & 70.0 & 80.0 & 10.0 & 53.3 & 60.0 & \underline{80.0} & 70.0 & \underline{70.0} & \underline{100.0} & \underline{100.0} & \underline{100.0} & \underline{100.0} & 92.2 & 84.4 & 74.4 \\
\textbf{C7 (Ours)} & 80.0 & \underline{90.0} & 60.0 & \underline{76.7} & 60.0 & \underline{80.0} & 70.0 & \underline{70.0} & \underline{100.0} & \underline{100.0} & \underline{100.0} & \underline{100.0} & \underline{94.4} & \underline{90.0} & \underline{82.2} \\
C8 & 80.0 & 80.0 & \underline{70.0} & \underline{76.7} & 30.0 & 50.0 & 70.0 & 50.0 & \underline{100.0} & \underline{100.0} & \underline{100.0} & \underline{100.0} & 86.7 & 85.6 & 75.6 \\
\bottomrule
\end{tabular}
\par\smallskip
\begin{minipage}{\linewidth}
\footnotesize
G/F/S: gentle/firm/strong; $n=10$ per prompt, $30$ per task, and $90$ overall. Task: completion within a command-associated attempt; + band: Task and pooled intent; Joint additionally requires pooled mean-force agreement. Underlines mark the best result in each column, including ties. Variant settings are given in Table~\ref{tab:vla_nominal_results}.
\end{minipage}
\end{table}

\begin{table}[!htbp]
\centering
\caption{Conditional force-tracking errors (N) over predicted-command activation windows. MAE pools absolute errors over timesteps; Mean averages per-episode mean-force errors. $n$ is the number of contributing episodes.}
\label{tab:vla_tracking_errors}
\label{tab:vla_pointwise_force_error}
\small
\setlength{\tabcolsep}{5pt}
\renewcommand{\arraystretch}{1.12}
\begin{tabular}{@{}l*{4}{rrr}@{}}
\toprule
& \multicolumn{3}{c}{Box pickup} & \multicolumn{3}{c}{Shelf pushing}
& \multicolumn{3}{c}{Surface wiping} & \multicolumn{3}{c}{Overall} \\
\cmidrule(lr){2-4}\cmidrule(lr){5-7}\cmidrule(lr){8-10}\cmidrule(lr){11-13}
Variant & MAE & Mean & $n$ & MAE & Mean & $n$ & MAE & Mean & $n$ & MAE & Mean & $n$ \\
\midrule
C0 & 1.75 & 2.17 & 28/30 & 6.06 & 4.78 & 30/30 & 3.13 & 2.52 & 30/30 & 3.65 & 3.16 & 88/90 \\
C1 & 2.04 & 1.65 & 28/30 & 4.28 & \underline{3.08} & 28/30 & 2.84 & 2.23 & 30/30 & 3.05 & 2.32 & 86/90 \\
C2 & 2.25 & 1.88 & 30/30 & 4.83 & 4.04 & 29/30 & 3.20 & 2.83 & 30/30 & 3.43 & 2.92 & 89/90 \\
C3 & 1.64 & 1.62 & 25/30 & 4.74 & 3.53 & 21/30 & 2.91 & 2.39 & 30/30 & 3.10 & 2.51 & 76/90 \\
C4 & 3.26 & 2.59 & 28/30 & 5.35 & 4.45 & 30/30 & 2.44 & 1.83 & 30/30 & 3.68 & 2.96 & 88/90 \\
C5 & 2.67 & 2.20 & 15/30 & 4.22 & 3.12 & 29/30 & 2.21 & 1.62 & 30/30 & 3.03 & 2.31 & 74/90 \\
C6 & 2.38 & 1.55 & 30/30 & 4.51 & 3.64 & 30/30 & 2.11 & 0.96 & 30/30 & 3.00 & 2.05 & 90/90 \\
\textbf{C7 (Ours)} & 2.08 & 1.45 & 30/30 & \underline{4.09} & 3.13 & 30/30 & \underline{2.04} & \underline{0.92} & 30/30 & \underline{2.74} & \underline{1.83} & 90/90 \\
C8 & \underline{1.55} & \underline{0.98} & 30/30 & 5.19 & 4.35 & 30/30 & 2.36 & 1.45 & 30/30 & 3.03 & 2.26 & 90/90 \\
\bottomrule
\end{tabular}
\par\smallskip
\begin{minipage}{\linewidth}
\footnotesize
Pickup averages hand-wise absolute errors; shelf pushing and wiping use the right hand. Overall equally weights tasks. Underlines mark the lowest error in each column, including ties. Episodes without command windows are excluded from error averages, not assigned zero, and remain failures in SR. Coverage differs, so these conditional errors alone do not rank policy performance.
\end{minipage}
\end{table}

\noindent\textbf{Force-tracking errors across variants.}
We evaluate pointwise force-tracking Mean Absolute Error (MAE) over the command-active windows used for scoring. Absolute differences between measured and VLA-issued forces are pooled across all included controller-rate samples, weighting longer command-active durations more heavily. For the Pickup task, we average the separately computed left- and right-hand MAEs; all other tasks evaluate the right hand only. The overall MAE is an unweighted average across the three tasks.

We also report episode mean-force error: for each hand, we take the absolute difference between commanded and measured force means pooled over all predicted-command activation windows, then average the hand-wise errors for pickup and average contributing episodes with equal episode weight. Task summaries include all three prompts, and overall values equally average the three task errors.

Table~\ref{tab:vla_tracking_errors} summarizes both task-level errors and contributing episode counts, while Figs.~\ref{fig:vla_mean_force_heatmap} and~\ref{fig:vla_force_error_heatmap} detail prompt-level results. Only episodes with predicted-command activation windows contribute to these error averages. Unsuccessful episodes contribute whenever these windows are present, regardless of prompt-band agreement or measured contact; measured zero-force samples within a window remain included. The $39$ episodes across C$0$--C$5$ that lack these windows have undefined errors and are excluded from both error averages, not assigned zero, though they remain failures in the success-rate denominators. All episodes in C$6$--C$8$ contain valid windows. Because the contributing episodes and issued commands differ across variants, these conditional errors alone do not rank policy performance. Pointwise MAE captures instantaneous discrepancies without allowing over- and undershoots to cancel, unlike the pooled mean-force agreement used by the tracking success gate. Note that low MAE does not guarantee correct force selection or task completion: a policy may accurately track an incorrectly low command.

For C7, the overall pointwise MAE and episode mean-force error are $2.74$~N and $1.83$~N, respectively, with all $90$ episodes contributing. Its shelf-pushing pointwise error increases from $3.57$~N under gentle prompts to $4.90$~N under strong prompts, while the corresponding mean-force error decreases from $3.46$~N to $2.89$~N. The two summaries also differ in timestep versus episode weighting.

For the default Opt2VLA configuration (C7), we additionally analyze issued force, measured force, and end-effector tracking error. Each episode contributes its mean over the same union of command-active windows used for force scoring; bars and whiskers show the mean and sample standard deviation across ten scenes per prompt. All C7 episodes have command-active windows and contribute regardless of task success or measured contact.

End-effector error is defined as the Euclidean distance between the VLA-issued controller target and the measured hand-body origin in the base-relative reset-heading frame. The instantaneous error is averaged within each episode before aggregation. These statistics describe command--execution discrepancies without uniquely attributing them to either the VLA policy or the low-level controller.

\noindent\textbf{C7 tracking analysis.}
Figure~\ref{fig:c7_tracking} shows force and end-effector tracking for the default Opt2VLA configuration. The largest difference between the plotted issued and measured force means is $2.69$~N for gentle shelf-box pushing, where the measured mean is below the issued mean. Across prompt/hand groups, mean end-effector errors span $3.22$--$4.66$~cm for box pickup, $7.70$--$8.77$~cm for shelf-box pushing, and $5.48$--$6.31$~cm for surface wiping. These command-window statistics include unsuccessful episodes; force agreement is scored per episode before aggregation across scenes.

\noindent\textbf{C7 failure breakdown.}
The Sankey diagram partitions each task's $30$ episodes according to the first unmet requirement: the task gate, pooled prompt-band agreement, or pooled mean-force agreement. Missing command-active windows remain failures. Ribbon widths represent episode counts rather than timesteps. These ordered gate failures describe observed conditions but do not uniquely attribute them to the VLA policy or low-level controller.

For box pickup, $5$ episodes fail the task gate and $2$ additional episodes fail prompt-band agreement. For shelf-box pushing, all episodes pass the task gate; $2$ fail prompt-band agreement and $7$ additional episodes fail mean-force agreement. Surface wiping passes all three criteria in all $30$ episodes.

\begin{figure}[!htbp]
\centering
\includegraphics[width=\textwidth]{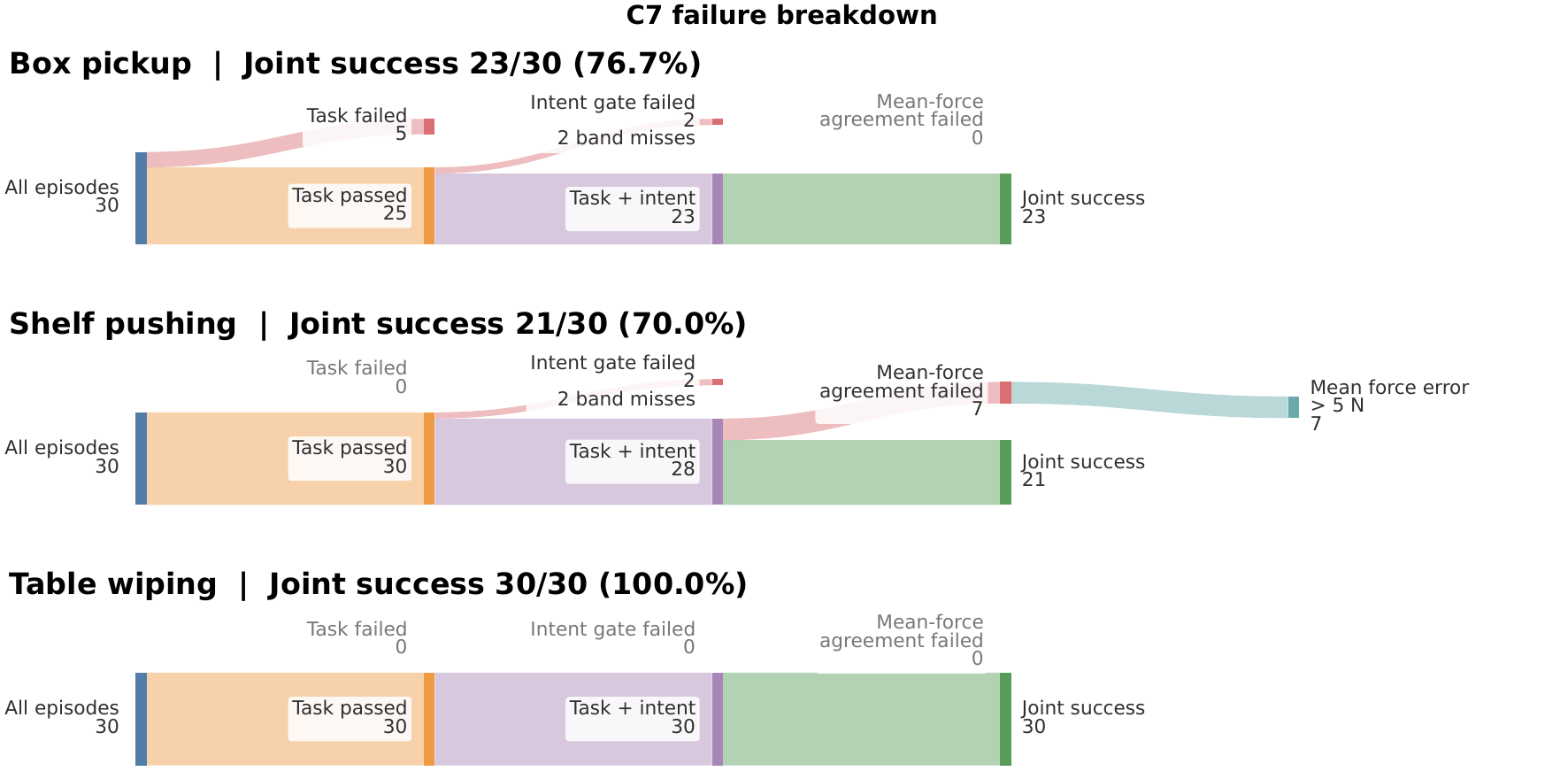}
\caption{\textbf{C7 failure breakdown.} Each task includes $30$ episodes, partitioned by the first unmet criterion. Joint success requires the task gate and both episode-level force gates.}
\label{fig:c7_failure_attribution}
\end{figure}
\begin{figure}[!htbp]
\centering
\includegraphics[width=\textwidth]{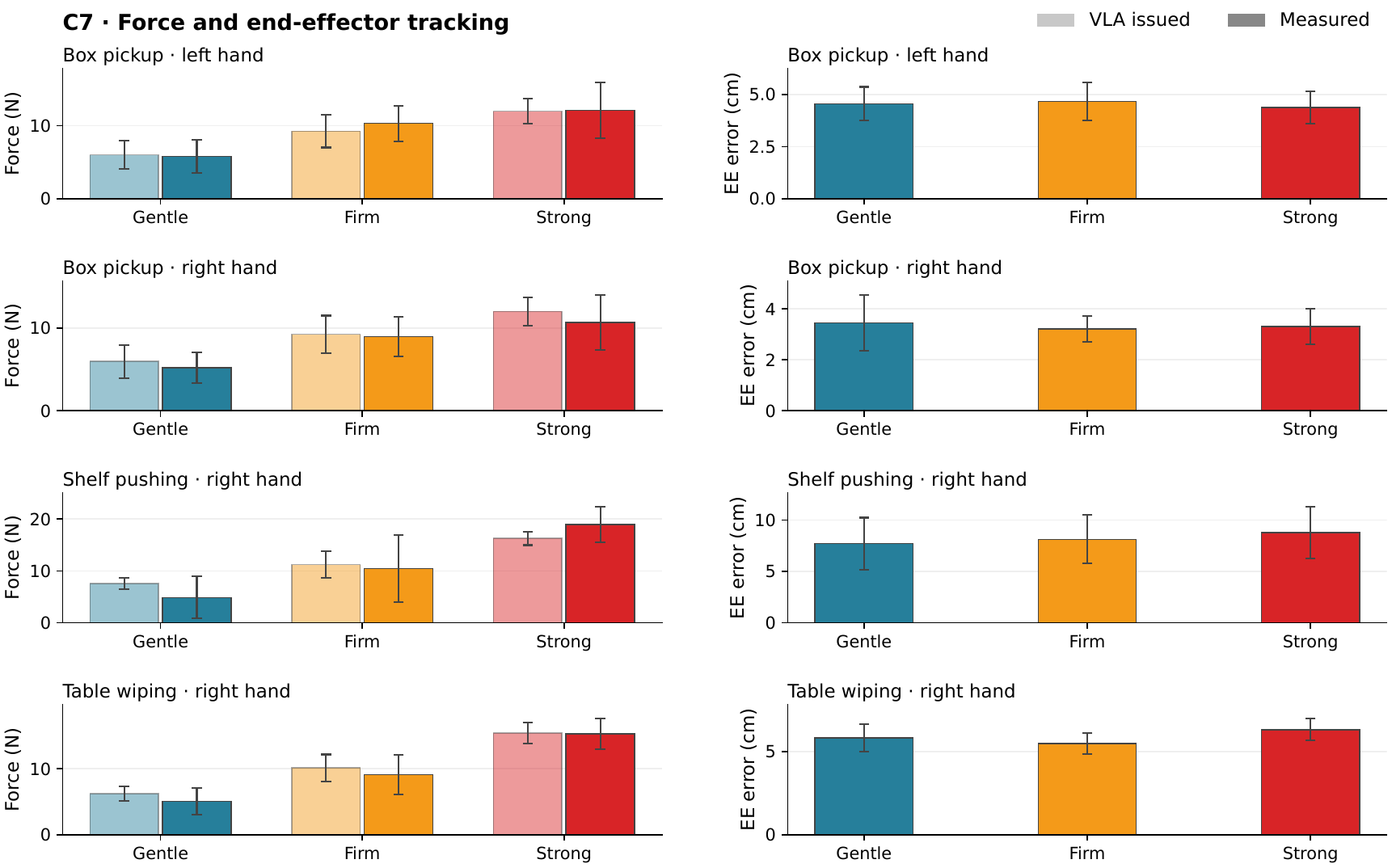}
\caption{\textbf{C7 tracking diagnostics.} Issued and measured force (left) and end-effector position error (right). Bars show means of episode means; whiskers show one sample standard deviation across ten scenes per prompt. All command-active intervals are included, regardless of task success or measured contact.}
\label{fig:c7_tracking}
\end{figure}
\begin{figure}[!htbp]
\centering
\includegraphics[width=\textwidth]{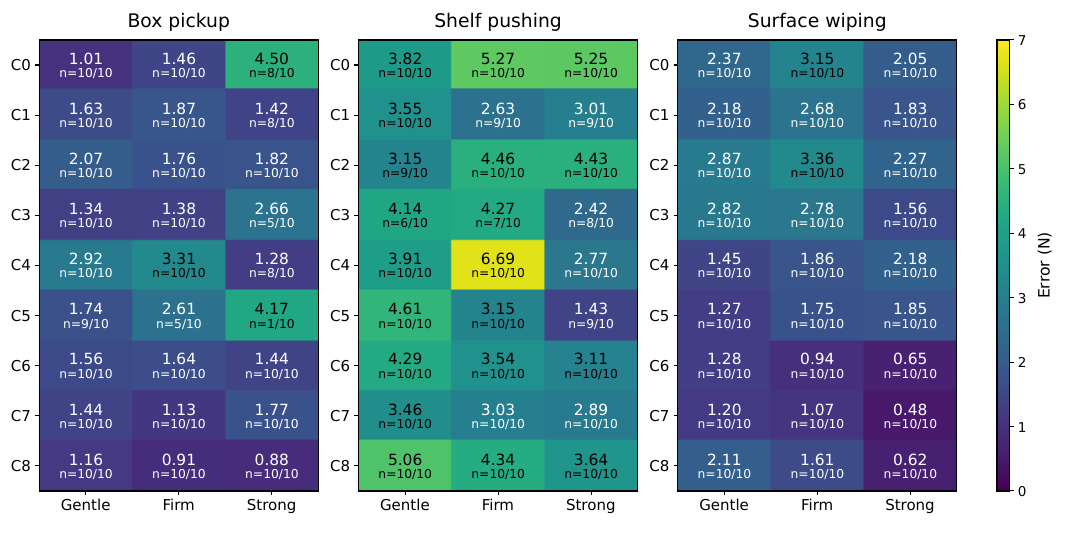}
\caption{\textbf{Episode mean-force error (N).} Each cell averages the per-episode absolute difference between commanded and measured means over all predicted-command activation windows. Pickup averages the two hand-wise errors. Counts show contributing episodes out of ten; episodes without command windows are excluded from these averages but remain failures in SR.}
\label{fig:vla_mean_force_heatmap}
\end{figure}
\begin{figure}[!htbp]
\centering
\includegraphics[width=\textwidth]{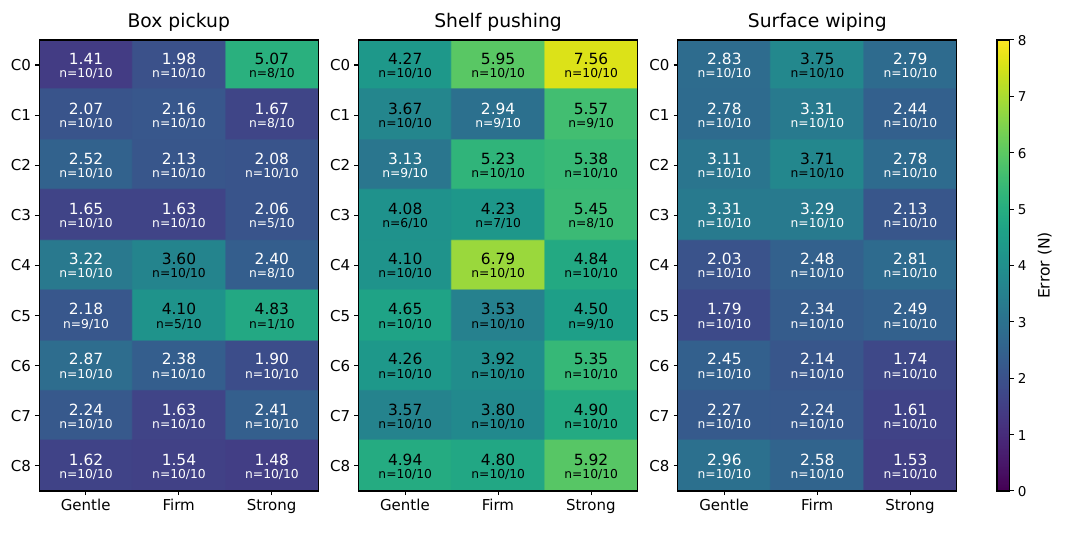}
\caption{\textbf{Pointwise force-tracking MAE by variant, task, and force prompt.} Values pool all command-active timesteps; pickup averages the two hand errors. Each cell reports MAE in newtons and the number of contributing episodes out of ten. Episodes without command-active windows are excluded from error aggregation but remain failures in success-rate denominators. The different timestep and episode weightings mean that the two heatmaps should not be interpreted as differing only in the order of averaging and taking absolute differences.}
\label{fig:vla_force_error_heatmap}
\label{fig:vla_pointwise_heatmap}
\end{figure}
\clearpage
\endgroup

\end{document}